\documentclass{llncs}
\usepackage{graphicx}
\usepackage{amstext}
\graphicspath{{Figures/}}    
\usepackage{hyperref}

\begin{document}

\title{FPD-M-net: Fingerprint Image Denoising and Inpainting Using M-Net Based Convolutional Neural Networks}

\titlerunning{FPD-M-net: Fingerprint Image Denoising and Inpainting Using M-net}
\author{Sukesh Adiga V \and Jayanthi Sivaswamy}
\authorrunning{Adiga et al.} 
\tocauthor{Jayanthi Sivaswamy}
\institute{Center for Visual Information Technology (CVIT), IIIT-Hyderabad, India \email{sukesh.adigav@research.iiit.ac.in}}

\maketitle              
\begin{abstract}
Fingerprint is a common biometric used for authentication and verification of an individual. These images are degraded when fingers are wet, dirty, dry or wounded and due to the failure of the sensors, etc. The extraction of the fingerprint from a degraded image requires denoising and inpainting. We propose to address these problems with an end-to-end trainable Convolutional Neural Network based architecture called \textit{FPD-M-net}, by posing the fingerprint denoising and inpainting problem as a segmentation (foreground) task. Our architecture is based on the \textit{M-net} with a change: structure similarity loss function, used for better extraction of the fingerprint from the noisy background. Our method outperforms the baseline method and achieves an overall 3rd rank in the \textit{Chalearn LAP Inpainting Competition Track 3$ - $Fingerprint Denoising and Inpainting, ECCV 2018}.


\end{abstract}


\section{Introduction}
\label{sec:introduction}
Fingerprint is an impression left by friction ridges of a finger. Human fingerprints are detailed, nearly unique, difficult to alter, and durable over the life of an individual, making them suitable as long-term biometrics for identifying the uniqueness of an individual. It plays an increasingly important role in security, to ensure privacy and identity verification. Fingerprint-based authentication is ubiquitous in day to day life (Example: unlocking in smartphones, mobile payments, international travel, accessing the restricted area, etc.). In forensic applications, the accuracy of fingerprint retrieval and verification systems are critical. However, recovery of fingerprints deposited on surfaces such as glass or metal or polished stone remains challenging.

Fingerprints details can be degraded due to impression conditions such as humidity, wet, dirty, skin dryness, and non-uniform contact with fingerprint capture device \cite{jain1997identity}. This results in poor image quality, hence require a denoising fingerprint information from the noise. In some cases, image can have missing regions due to the failure of fingerprint sensors or wound in finger. It requires a filling or inpainting from the neighbouring region. Overall, fingerprint image denoising and inpainting can be seen as a preprocessing step to ease subsequent operations like fingerprint authentication and verification carried out either by humans or existing software.

There are many methods for fingerprint enhancement in literature. Early efforts were based on traditional image filtering methods with a directional median filter \cite{wu2004fingerprint}, Wiener filter and anisotropic filter \cite{greenberg2002fingerprint}. A partial differential equation \cite{rahmes2007fingerprint} based method was proposed for automated fingerprint reconstruction. Several methods use orientations information to enhance fingerprint quality. Hong et al. \cite{hong1998fingerprint} use ridge orientation and frequency information to improve the clarity of ridge and valley structures in fingerprint image. \cite{singh2015fingerprint}. Feng et al. \cite{feng2013orientation} and Yang et al. \cite{yang2014localized} proposed a dictionary approach for orientation estimation to improve latent fingerprint. Chen et al. \cite{chen2016multi} used multiscale dictionaries to handle a varying level of noise in fingerprint image.

Recently, Convolution Neural Networks (CNN) have been successful in many computer vision tasks such as segmentation, denoising, and inpainting. Some of the recent works are explored using CNN for fingerprint extraction and analysis. Sahasrabudhe et al. \cite{sahasrabudhe2014fingerprint} use a deep belief network to learn features from greyscale to clean fingerprint images. Cao et al. \cite{cao2015latent} pose latent orientation estimation as a patch classification problem using CNN. Tang et al. \cite{tang2017fingernet} proposed a FingerNet, based on deep convolutional network. It uses domain knowledge for fingerprint minutiae extraction in noisy ridge patterns and complex background. The network first segment the orientation field, then it enhances latent fingerprint to obtain minutiae. Recently, Li et al. \cite{li2018deep} developed a method based on FingerNet to enhance the fingerprint images. Nguyen et al. \cite{nguyen2018robust} proposed a network called MinutiaeNet, consists of course and fine network which does a fully automatic minutiae extraction. Here, course network uses domain knowledge to enhance an image and extract segmentation map to give candidate minutiae locations. Fine network refines the candidate minutiae locations. Another interesting approach is based on the generative network to improve fingerprint images. Svoboda et al. \cite{svoboda2017generative} proposed a generative convolutional network to denoise and predict the missing parts of the ridge pattern in latent fingerprint image.

Success of deep learning in dealing with inpainting and denoising \cite{xie2012image} problems has led to the ChaLearn competition\footnote{http://chalearnlap.cvc.uab.es/challenge/26/track/32/description/} \cite{sergio2019denoising} which focuses on the development of a deep learning solution to restore fingerprint images from the degraded images. In our work, we pose a given problem as segmenting fingerprint from the noisy background and hence propose a solution using an architecture developed for object segmentation.

\section{Method}
\label{sec:method}
The distorted fingerprint images require denoising and inpainting for the restoration of accurate ridges which helps in reliable authentication and verification. The image consists of an object of interest (i.e., fingerprint) in a noisy or cluttered background. The problem can be solved using segmentation of object (fingerprint) from the noisy background. The M-net \cite{mehta2017m} does excellent segmentation, which forms motivation for our work.

Our aim is to denoise and inpaint the fingerprint images simultaneously using a segmentation approach, where fingerprint information is foreground of interest and other details are background. The filling of any missing information should be possible with appropriate training, rather than explicit inpainting. The M-net was proposed for 3D brain structure segmentation, where an initial block converts 3D information into a 2D image on which segmentation is performed. Further, a categorical cross entropy loss function is used for segmentation. The 3D-to-2D conversion block is redundant, hence dropped and the loss function is also changed to suit the task at hand. The resulting architecture is called \textit{FPD-M-net}. The details of network architecture, training and loss function are described next.

\subsection{FPD-M-net architecture}
\label{ssec:arch}
The U-net \cite{ronneberger2015u} architecture is commonly used for tasks such as segmentation or restoration. The M-net is modified U-net for better segmentation. It uses 3D information for segmentation; hence a 3D-to-2D converter block is introduced. M-net also has four pathways to have similar functionality as deep-supervision \cite{lee2015deeply}. It introduces two side paths (left and right leg) along with two main encoding and decoding paths. The left leg path, downsample the input and given to corresponding encoder layers. The right leg path, upsample the output of each of the decoding layers to original size. The final layer combines the outputs of right leg and decoder layer to give a final output.

\begin{figure}[t]
    \centering
    \includegraphics[width=1\textwidth]{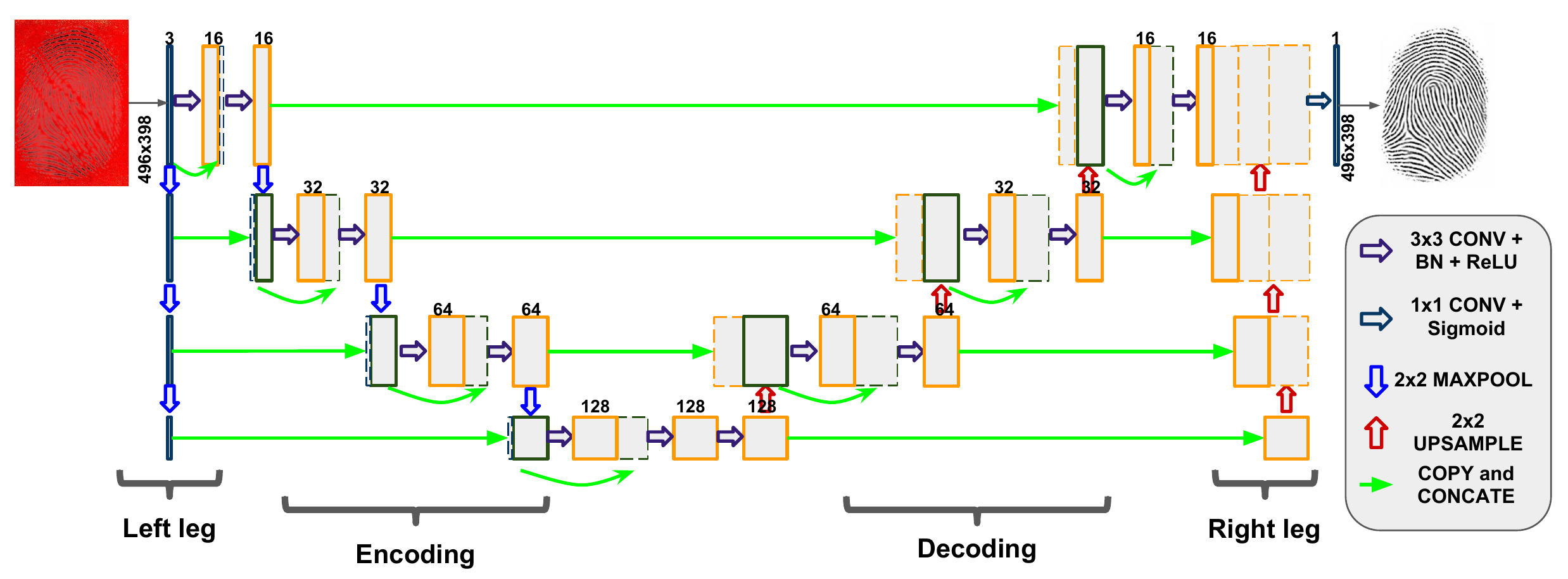}
    \caption{The schematic representation of FPD-M-net architecture. Solid yellow boxes represent the output of CONV-BN-ReLU block. Dashed boxes represent copied feature maps. The number of feature maps is denoted on top of the box}
    \label{fig:FPD-M-net}
\end{figure}

Our FPD-M-net architecture is adapted from M-net \cite{mehta2017m}. It consists of a Convolutional layer (CONV), maxpooling layer, upsampling layer, Dropout layer \cite{srivastava2014dropout}, Batch Normalisation layer (BN) \cite{ioffe2015batch}, and rectified linear unit (ReLU) activation functions with encoder and decoder style of architecture as shown in Fig.~\ref{fig:FPD-M-net}. Encoding layer consists of repeated two blocks of $3 \times 3$ CONV, BN, and ReLU. Between two blocks of CONV-BN-ReLU layer, a dropout layer (with probability 0.2) is included. Dropout layer prevents over-fitting, BN layer enables faster and more stable training. The output of two blocks CONV-BN-ReLU are concatenated and downsampled with a $2 \times 2$ maxpooling operation with stride 2. Decoder layer is similar to encoder layer with one exception: maxpooling is replaced by upsampling layer which helps to reconstruct an output image. The final layer is a $1 \times 1$ convolution layer with a sigmoid activation function which gives the reconstructed output image.

Skip connections used in FPD-M-net are shown (with green arrows) in Fig.~\ref{fig:FPD-M-net}. The skip connection between adjacent convolution filters, enables the network to learn better features \cite{srivastava2015highway} and the skip connection from input-to-encoder (left leg), encoder-to-decoder, and decoder-to-output (right leg) ensures that network has sufficient information to drives fine grain details of fingerprint image. There are some differences between FPD-M-net and M-net, which helped the task at hand. Differences are as follows: i) Conv-ReLU-BN blocks are replaced with Conv-BN-ReLU blocks as in BN paper \cite{ioffe2015batch} (See section \ref{sssec:BN}); ii) a combination of a per-pixel loss and structure similarity loss are used for loss function as the ground-truth fingerprint image is integer valued in the range [0, 255]; iii) in final layer, sigmoid activation function instead of softmax activation to obtain output image, as our task here is to reconstruct fingerprint image.

\subsection{Training details}
\label{ssec:training}
Network is trained end-to-end with the pair of noisy/distorted and clean/ground-truth fingerprint image. Input and ground-truth images are padded with edge values to suit the network and images are normalised to take values between [0,1]. The size of input and ground truth images are $275 \times 400$ pixels. After padding, size of images become $368 \times 496$. Padding is done so that output of the network effectively sees the input image size of $275 \times 400$. In testing phase, distorted images are given to FPD-M-net, to get a clean fingerprint image as output. The output images are unpadded to match original size and compared against reference image.

\subsection{Loss function}
\label{ssec:loss}
The mean squared error (MSE), a reference-based metric and Peak Signal-to-Noise Ratio (PSNR) are popular error measures for reconstruction problems. In deep learning, MSE is widely used as a loss function for many applications. However, neither MSE nor PSNR correlates well with human perception of image quality. Structure similarity index (SSIM) \cite{wang2004image} is a reference-based metric that has been developed for this purpose. The SSIM is measured at a fixed scale and may only be appropriate for a certain range of image scales. A more advanced form of SSIM is multi-scale structure similarity index (MS-SSIM) \cite{wang2003multi}. It preserves the structure and contrast in high-frequency regions better than other loss functions \cite{zhao2017loss}. In addition to choosing perceptually correlated metric, it is also of interest to preserve intensity as ground-truth fingerprint image has real value. So we choose a combination of per-pixel loss and MS-SSIM to define the loss function with weight $\delta$, as shown:
\begin{equation}
L(\theta) = \delta \cdot L_{\text{MS-SSIM}}(\theta) + (1-\delta) \cdot L_{l_1}(\theta)
\end{equation}
where, $L_{l_1}(\theta)$ is $l_1$ loss and $L_{\text{MS-SSIM}}(\theta)$ is standard MS-SSIM loss. The weights are set to $\delta = 0.85$ as per \cite{zhao2017loss} and MS-SSIM is computed over three scales.

\section{Experiments and Results}
\subsection{Dataset and Parameters}
\label{ssec:dataset}
Dataset used in our experiment is obtained with the \textit{Anguli: Synthetic Fingerprint Generator} software, provided by the \textit{Chalearn LAP Inpainting Competition Track 3\footnote{http://chalearnlap.cvc.uab.es/dataset/32/description/}}. Dataset consists of a pair of degraded/distorted and ground-truth fingerprint images. The distorted images are synthetically generated by first degrading fingerprints with a distortion model which introduces blur, brightness, contrast, elastic transformation, occlusion, scratch, resolution, rotation and then overlaying fingerprints on top of various backgrounds. The dataset consists of training, validation and test sets with a pair of degraded and ground-truth fingerprint images. It is described in Table~\ref{table:dataset}. The images are padded and normalised before training and testing. Test set has no ground-truth and evaluation requires uploading the images to the competition site to get a quantitative score.

\begin{table}[h!]
\centering
\addtolength{\tabcolsep}{20pt}
\begin{tabular}{ c | c}
    \hline \hline
    Dataset      & Number of images \\ [0.5ex] \hline \hline
    Training     & 75,600           \\ 
    Validation   & 8,400            \\ 
    Test         & 8,400            \\ [0.3ex] \hline \hline
\end{tabular}
\caption{Fingerprint images dataset}
\label{table:dataset}
\end{table}

The FPD-M-net was trained for 75 epochs for a week. A stochastic gradient descent (SGD) optimiser was used to minimise the per-pixel loss and structure similarity loss. The training parameters were: learning rate of 0.1; Nesterov momentum was set to 0.75; decay rate was set at 0.00001; batch size was chosen as 8. After 50 epochs learning rate was reduced to 0.01; Nesterov momentum was increased to 0.95. Network parameters are presented in Table~\ref{table:parameter}. Network was implemented on an NVIDIA GTX 1080 GPU, with 12GB of GPU RAM on a core i7 processor. The entire architecture was implemented in Keras library using Theano backend. Code of our method has been publicly released\footnote{https://github.com/adigasu/FDPMNet}.

\begin{table}[h!]
\centering
\addtolength{\tabcolsep}{10pt}
\begin{tabular}{ c | c | c}
    \hline \hline
    Parameter         & First 50 epoch & After 50 epoch \\ [0.5ex] \hline \hline
    Learning Rate     & 0.1            & 0.01           \\ 
    Nesterov momentum & 0.75           & 0.95           \\ 
    Decay rate        & 0.00001        & 0.00001        \\ 
    Batch size        & 8              & 8              \\ [0.3ex] \hline \hline
\end{tabular}
\caption{FPD-M-net training parameters}
\label{table:parameter}
\end{table}

\subsection{Results and performance evaluation}
\label{ssec:results}
The results of FPD-M-net were evaluated both qualitatively and quantitatively. We first compared it with U-net architecture using metrics such as PSNR, MSE. The perceptual quality of results was evaluated using structural similarity (SSIM). Next, we provide a performance comparison with other participants of \textit{Chalearn LAP Inpainting Competition Track 3$ - $Fingerprint Denoising and Inpainting, ECCV 2018}. Finally, sample qualitative results are presented.

\subsubsection{Performance evaluation with U-net}
\label{sssec:unet}
The quantitative comparison of results of FPD-M-net is compared against U-net (trained with the same setting as FPD-M-net). U-net was trained with the same loss function with only encoder-to-decoder skip connection \cite{ronneberger2015u}. The denoising and inpainting performance of fingerprint images are evaluated using PSNR, MSE and SSIM metric. These results are presented in Table~\ref{table:unetcmp} for both validation and test sets. Our method outperforms U-net in all metrics, which indicates skip connections aid in achieving superior fingerprint restoration.

\begin{table}[h!]
\centering
\addtolength{\tabcolsep}{10pt}
\begin{tabular}{ c | c | c | c | c }
    \hline \hline
    set        & method    & MSE    & PSNR    & SSIM   \\ \hline \hline
    validation & U-net     & 0.0286 & 16.2122 & 0.8202 \\
               & FPD-M-net & \textbf{0.0270} & \textbf{16.5149} & \textbf{0.8255} \\
    \hline
    test       & U-net     & 0.0284 & 16.2466 & 0.8207 \\
               & FPD-M-net & \textbf{0.0268} & \textbf{16.5534} & \textbf{0.8261} \\ \hline\hline
\end{tabular}
\caption{Quantitative comparison of results of FPD-M-net with U-net}
\label{table:unetcmp}
\end{table}

\subsubsection{Ablation experiments with batch normalisation}
\label{sssec:BN}
In order to assess effect of batch normalisation (BN) before and after the activation function, two FPD-M-net networks were trained: one with BN after ReLU activation (similar to M-net) and one with BN before ReLU activation. For convenience, BN after and before ReLU activation network called as FPD-M-net-A and FPD-M-net-B, respectively. Both the networks were trained with the same settings as described in section~\ref{ssec:dataset}. The quantitative results for validation and test set are presented in Table~\ref{table:mnetcmp}. Results indicate that FPD-M-net-B is better in PSNR and MSE metric than FPD-M-net-A, whereas for SSIM, FPD-M-net-A has slightly better than FPD-M-net-B. Overall FPD-M-net-B has a fairly good results relative to FPD-M-net-A. Hence, BN before ReLU activation function is used in FPD-M-net.

\begin{table}[h!]
\centering
\addtolength{\tabcolsep}{10pt}
\begin{tabular}{ c | c | c | c | c }
    \hline \hline
    set        & method    & MSE    & PSNR    & SSIM   \\ \hline \hline
    validation & FPD-M-net-B & \textbf{0.0270} & \textbf{16.5149} & 0.8255 \\
               & FPD-M-net-A & 0.0277 & 16.4019 & \textbf{0.8265} \\
    \hline
    test       & FPD-M-net-B & \textbf{0.0268} & \textbf{16.5534} & 0.8261 \\
               & FPD-M-net-A & 0.0275 & 16.4336 & \textbf{0.8270} \\ \hline\hline
\end{tabular}
\caption{Quantitative comparison of FPD-M-net with BN before and after activation function}
\label{table:mnetcmp}
\end{table}

\subsubsection{Comparison with others in Challenge}
\label{sssec:challenge}
Fingerprint denoising and inpainting challenge was organised by \textit{Chalearn LAP Inpainting Competition, ECCV 2018}. The final quantitative results of competition are presented in Table~\ref{table:chal}. The CVxTz and rgsl888 team also used a U-net \cite{ronneberger2015u} based architecture, whereas hcilab team used a hierarchical deep learning approach \cite{salakhutdinov2013learning}. The baseline network provided in competition is a standard deep neural network\footnote{http://chalearnlap.cvc.uab.es/challenge/26/track/32/baseline/} with residual blocks. The rgsl888 team uses a dilated convolutions compared to CVxTz team. In our U-net implementation (section~\ref{sssec:unet}), a combination of $l_1$ and MS-SSIM loss function is used whereas CVxTz and rgsl888 used $l_1$ and $l_2$ loss function, respectively. The overall CVxTz team performs the best. It should be noted that U-net network used by CVxTz team has almost double the network depth as compared to our FPD-M-net and also used additional data augmentation. Our method obtains $0.8261$ (rank 2) in SSIM metric, which shows the effectiveness of MS-SSIM in loss function.

\begin{table}[ht!]
\centering
\addtolength{\tabcolsep}{8pt}
\begin{tabular}{ c | c | c | c | c }
    \hline \hline
    Team          & Rank    & MSE        & PSNR        & SSIM   \\ \hline \hline
    CVxTz        & 1.0000  & 0.0189 (1) & 17.6968 (1) & 0.8427 (1) \\
    rgsl888      & 2.3333  & 0.0231 (2) & 16.9688 (2) & 0.8093 (3) \\
    sukeshadigav (FPD-M-net) & 3.3333  & 0.0268 (4) & 16.5534 (4) & 0.8261 (2) \\
    hcilab       & 3.3333  & 0.0238 (3) & 16.6465 (3) & 0.8033 (4) \\
    Baseline 	 & 4.6666  & 0.0241 (5)	& 16.4109 (5) & 0.7965 (5) \\
    \hline\hline
\end{tabular}
\caption{Performance of different methods in the Challenge}
\label{table:chal}
\end{table}

\subsubsection{Qualitative results}
\label{sssec:qualitative}
\begin{figure*}[t!]
    \centering
    \includegraphics[width=1.0\textwidth]{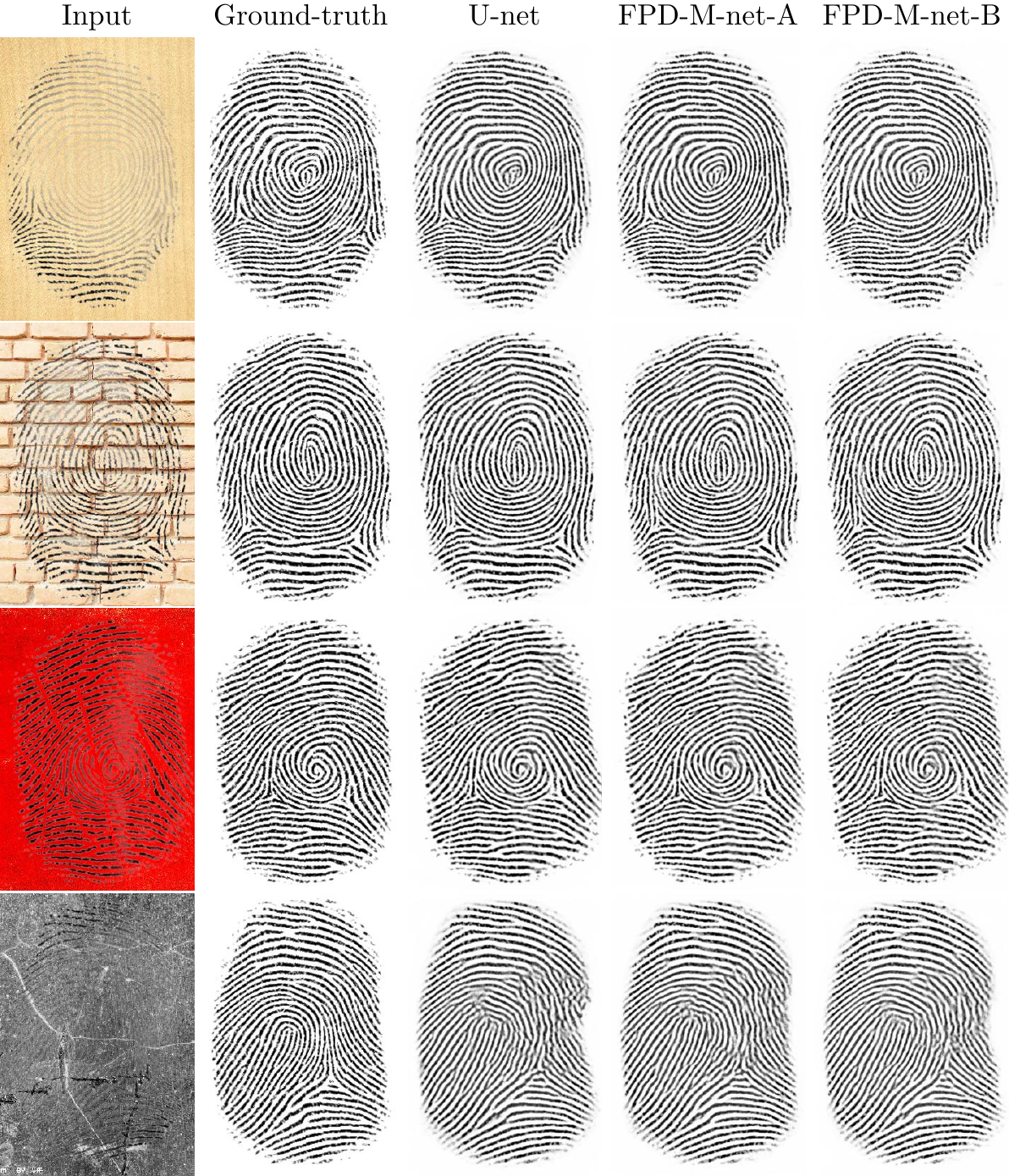}
    \caption{Illustration of fingerprint denoising and inpainting results for varying distorted images. From left to right: distorted fingerprints, corresponding ground-truth, results of U-net, our methods FPD-M-net-A and FPD-M-net-B}
    \label{fig:res}
\end{figure*}

A qualitative comparison of fingerprint image denoising and inpainting can be done with sample images from the test set which are shown in Fig.~\ref{fig:res}. Two moderately distorted (Row 1 and 2) and two severely distorted fingerprint images (Row 3 and 4) and its corresponding results are shown. The weak fingerprints are successfully recovered as shown in Row 1. Networks are robust to even strong background clutter (Row 2). Automatic filling is seen to be successful in images in Row 3 and 4. Our FPD-M-net method produces better results for severely distorted images (Row 4) compared to U-net.

\begin{figure*}[t!]
    \centering
    \includegraphics[width=1.0\textwidth]{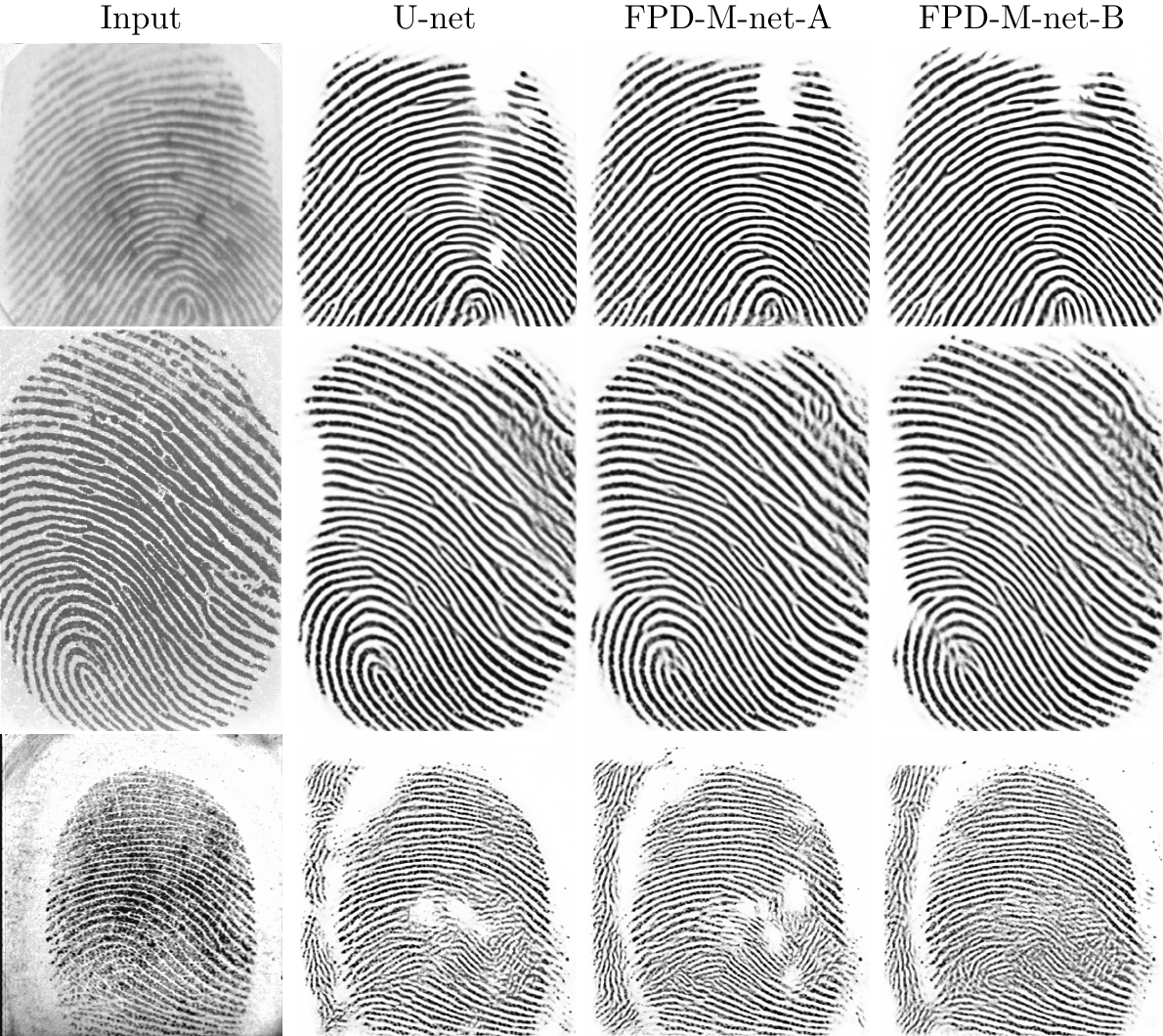}
    \caption{Sample results of fingerprint denoising and inpainting on real images. From left to right: distorted fingerprints, results of U-net, our methods FPD-M-net-A and FPD-M-net-B}
    \label{fig:res2}
\end{figure*}

\paragraph{\textbf{Qualitative comparison with real fingerprint}}
Since images provided in the Challenge were synthetically generated it is of interest to test the proposed architecture on real images also. The qualitative performance of denoising and inpainting results on real images from three datasets: FVC2000 DB1, DB2 and DB3 \cite{maio2002fvc2000} are shown in Fig.~\ref{fig:res2}. These datasets are captured by different sensors having varying resolutions. DB1 images appear closer to synthetic dataset. A sample image from DB1 (Row 1), DB2 (Row 2) and DB3 (Row 3) along with outputs are shown in Fig.~\ref{fig:res2}. The FPD-M-net methods produce the better result for DB1 image compared to U-net. In case of a DB2 image, portions fingerprint are missing in top and left part of the image. Some artefact is also seen in all the results in top right of the image. Apart from these defects, all methods perform fairly well. In case of a DB3 image, all results exhibit some loss of information, unlike FPD-M-net-B which however has some distortion (in the lower part). The difference in the results of testing on synthetic versus real images could be due to a number of factors including variation in acquisition (sensors and resolutions) which affect the width of ridges.

\section{Conclusion}
In this work, we presented an FPD-M-net model for fingerprint denoising and inpainting using a pair of synthetic data. The segmentation based architecture is shown to handle both denoising and inpainting of fingerprint images, simultaneously. It outperforms the U-net, and baseline model which is given in the competition. Our model is robust to strong background clutter, weak signal and performs automatic filling effectively. Perceptual results for both qualitatively and quantitatively indicate the effectiveness of the MS-SSIM loss function. Results for images acquired with different sensors suggest the need for sensor-specific training for better results.

\bibliographystyle{splncs03}
\bibliography{fpdMNet}

\begin{thebibliography}{10}
\providecommand{\url}[1]{\texttt{#1}}
\providecommand{\urlprefix}{URL }

\bibitem{cao2015latent}
Cao, K., Jain, A.K.: Latent orientation field estimation via convolutional
  neural network. In: 2015 International Conference on Biometrics (ICB). pp.
  349--356. IEEE (2015)

\bibitem{chen2016multi}
Chen, C., Feng, J., Zhou, J.: Multi-scale dictionaries based fingerprint
  orientation field estimation. In: 2016 International Conference on Biometrics
  (ICB). pp. 1--8. IEEE (2016)

\bibitem{feng2013orientation}
Feng, J., Zhou, J., Jain, A.K.: Orientation field estimation for latent
  fingerprint enhancement. IEEE transactions on pattern analysis and machine
  intelligence  35(4),  925--940 (2013)

\bibitem{greenberg2002fingerprint}
Greenberg, S., Aladjem, M., Kogan, D.: Fingerprint image enhancement using
  filtering techniques. Real-time Imaging  8(3),  227--236 (2002)

\bibitem{hong1998fingerprint}
Hong, L., Wan, Y., Jain, A.: Fingerprint image enhancement: Algorithm and
  performance evaluation. IEEE Transactions on pattern analysis and machine
  intelligence  20(8),  777--789 (1998)

\bibitem{ioffe2015batch}
Ioffe, S., Szegedy, C.: Batch normalization: Accelerating deep network training
  by reducing internal covariate shift. arXiv preprint arXiv:1502.03167  (2015)

\bibitem{jain1997identity}
Jain, A.K., Hong, L., Pankanti, S., Bolle, R.: An identity-authentication
  system using fingerprints. Proceedings of the IEEE  85(9),  1365--1388 (1997)

\bibitem{lee2015deeply}
Lee, C.Y., Xie, S., Gallagher, P., Zhang, Z., Tu, Z.: Deeply-supervised nets.
  In: Artificial Intelligence and Statistics. pp. 562--570 (2015)

\bibitem{li2018deep}
Li, J., Feng, J., Kuo, C.C.J.: Deep convolutional neural network for latent
  fingerprint enhancement. Signal Processing: Image Communication  60,  52--63
  (2018)

\bibitem{maio2002fvc2000}
Maio, D., Maltoni, D., Cappelli, R., Wayman, J.L., Jain, A.K.: Fvc2000:
  Fingerprint verification competition. IEEE Transactions on Pattern Analysis
  \& Machine Intelligence (3),  402--412 (2002)

\bibitem{mehta2017m}
Mehta, R., Sivaswamy, J.: M-net: A convolutional neural network for deep brain
  structure segmentation. In: Proc. of 14th International Symposium on
  Biomedical Imaging (ISBI). pp. 437--440. IEEE (2017)

\bibitem{nguyen2018robust}
Nguyen, D.L., Cao, K., Jain, A.K.: Robust minutiae extractor: Integrating deep
  networks and fingerprint domain knowledge. In: 2018 International Conference
  on Biometrics (ICB). pp. 9--16. IEEE (2018)

\bibitem{rahmes2007fingerprint}
Rahmes, M., Allen, J.D., Elharti, A., Tenali, G.B.: Fingerprint reconstruction
  method using partial differential equation and exemplar-based inpainting
  methods. In: Biometrics Symposium. pp. 1--6. IEEE (2007)

\bibitem{ronneberger2015u}
Ronneberger, O., Fischer, P., Brox, T.: U-net: Convolutional networks for
  biomedical image segmentation. In: International Conference on Medical image
  computing and computer-assisted intervention (MICCAI). pp. 234--241. Springer
  (2015)

\bibitem{sahasrabudhe2014fingerprint}
Sahasrabudhe, M., Namboodiri, A.M.: Fingerprint enhancement using unsupervised
  hierarchical feature learning. In: Proceedings of Indian Conference on
  Computer Vision Graphics and Image Processing. p.~2. ACM (2014)

\bibitem{salakhutdinov2013learning}
Salakhutdinov, R., Tenenbaum, J.B., Torralba, A.: Learning with
  hierarchical-deep models. IEEE transactions on pattern analysis and machine
  intelligence  35(8),  1958--1971 (2013)

\bibitem{sergio2019denoising}
Sergio, E., et~al.: Chalearn looking at people: Inpainting and denoising
  challenges. Challenges in Machine Learning (CiML)  (2019)

\bibitem{singh2015fingerprint}
Singh, K., Kapoor, R., Nayar, R.: Fingerprint denoising using ridge orientation
  based clustered dictionaries. Neurocomputing  167,  418--423 (2015)

\bibitem{srivastava2014dropout}
Srivastava, N., Hinton, G., Krizhevsky, A., Sutskever, I., Salakhutdinov, R.:
  Dropout: a simple way to prevent neural networks from overfitting. The
  Journal of Machine Learning Research  15(1),  1929--1958 (2014)

\bibitem{srivastava2015highway}
Srivastava, R.K., Greff, K., Schmidhuber, J.: Highway networks. arXiv preprint
  arXiv:1505.00387  (2015)

\bibitem{svoboda2017generative}
Svoboda, J., Monti, F., Bronstein, M.M.: Generative convolutional networks for
  latent fingerprint reconstruction. In: 2017 IEEE International Joint
  Conference on Biometrics (IJCB). pp. 429--436. IEEE (2017)

\bibitem{tang2017fingernet}
Tang, Y., Gao, F., Feng, J., Liu, Y.: Fingernet: An unified deep network for
  fingerprint minutiae extraction. In: IEEE International Joint Conference on
  Biometrics (IJCB). pp. 108--116. IEEE (2017)

\bibitem{wang2004image}
Wang, Z., Bovik, A.C., Sheikh, H.R., Simoncelli, E.P.: Image quality
  assessment: from error visibility to structural similarity. IEEE Transactions
  on Image Processing  13(4),  600--612 (2004)

\bibitem{wang2003multi}
Wang, Z., Simoncelli, E., Bovik, A., et~al.: Multi-scale structural similarity
  for image quality assessment. In: ASILOMAR CONFERENCE ON SIGNALS SYSTEMS AND
  COMPUTERS. vol.~2, pp. 1398--1402. IEEE (2003)

\bibitem{wu2004fingerprint}
Wu, C., Shi, Z., Govindaraju, V.: Fingerprint image enhancement method using
  directional median filter. In: Biometric Technology for Human Identification.
  vol. 5404, pp. 66--76. International Society for Optics and Photonics (2004)

\bibitem{xie2012image}
Xie, J., Xu, L., Chen, E.: Image denoising and inpainting with deep neural
  networks. In: Advances in neural information processing systems. pp. 341--349
  (2012)

\bibitem{yang2014localized}
Yang, X., Feng, J., Zhou, J.: Localized dictionaries based orientation field
  estimation for latent fingerprints. IEEE transactions on pattern analysis and
  machine intelligence  36(5),  955--969 (2014)

\bibitem{zhao2017loss}
Zhao, H., Gallo, O., Frosio, I., Kautz, J.: Loss functions for image
  restoration with neural networks. IEEE Transactions on Computational Imaging
  3(1),  47--57 (2017)

\end{thebibliography}

\end{document}